# Pearl's Calculus of Intervention Is Complete

**Yimin Huang** and **Marco Valtorta**
Department of Computer Science and Engineering
University of South Carolina
Columbia, South Carolina 29208
{huang6, mgv}@cse.sc.edu

## Abstract

This paper is concerned with graphical criteria that can be used to solve the problem of identifying casual effects from nonexperimental data in a causal Bayesian network structure, i.e., a directed acyclic graph that represents causal relationships. We first review Pearl's work on this topic [Pearl, 1995], in which several useful graphical criteria are presented. Then we present a complete algorithm [Huang and Valtorta, 2006b] for the identifiability problem. By exploiting the completeness of this algorithm, we prove that the three basic *do-calculus rules* that Pearl presents are complete, in the sense that, if a causal effect is identifiable, there exists a sequence of applications of the rules of the do-calculus that transforms the causal effect formula into a formula that only includes observational quantities.

## 1 Introduction

This paper focuses on graphical criteria used to infer the strength of cause-and-effect relationships from a causal Bayesian network [Pearl, 1995, Pearl, 2000], which is an acyclic directed graph representing nonexperimental data and causal relationships.

In the 1990s, some graphical conditions were given to show whether the causal effect, that is, the joint response of any set $S$ of variables to interventions on a set $T$ of action variables, denoted as $P_T(S)$[1], is identifiable or not. Those results are summarized in [Pearl, 2000]. For example, "back-door" and "front-door" criteria and do-calculus [Pearl, 1995]; graphical criteria to identify $P_T(S)$ when $T$ is a singleton [Galles and Pearl, 1995]; special graphical conditions under which it is possible to identify $P_T(S)$ [Pearl and Robins, 1995]. Some further study can be also found in [Robins, 1997] and [Kuroki and Miyakawa, 1999]. In all these graphical criteria, Pearl's three do-calculus (inference) rules are in the core position. All the other graphical rules can be obtained from them. Pearl conjectures that they are sufficient for the identification problem, but the conjecture has remained opened until now.

In the current decade, Tian and Pearl published a series of papers related to the identification problem [Tian and Pearl, 2002a, Tian and Pearl, 2002b, Tian and Pearl, 2003]. Their new methods combined the graphical character of causal graph and the algebraic definition of causal effect. They used both algebraic and graphical methods to identify causal effects.

Based on their work, Huang and Valtorta proved that Tian and Pearl's identify algorithm for semi-Markovian graphs is complete [Huang and Valtorta, 2006a]. Here, semi-Markovian graphs are defined as causal graphs in which each unobservable variable is a root and has exactly two observable children; semi-Markovian graphs are sometimes defined differently. It has been shown that a transformation between general Bayesian networks and semi-Markovian graphs, defined, e.g., in [Tian and Pearl, 2002b], preserves identifiability [Huang and Valtorta, 2006b]. In [Huang and Valtorta, 2006b], the authors also present an algorithm on general causal Bayesian networks and prove that the algorithm is complete, which means a causal effect is identifiable if and only if the given algorithm runs successfully and returns an expression that is the target causal effect in terms of estimable quantities.

We have recently learned that the sufficiency of the three inference rules (with some minor technical limitations) has been proved in [Shpitser and Pearl, 2006]. Our independently obtained result applies to general causal Bayesian networks.

In this paper, we review the graphical rules and the complete identify algorithm of [Huang and Valtorta, 2006b]. We consider their relationship and prove that the identify algorithm can be obtained by using the three inference rules. Because of the completeness of the identify algorithm, our proof shows that the three inference rules are sufficient, which confirms Pearl's conjecture. In the next section we give out the definitions and notations that we use in this paper. In section three, we review some graphical rules for identification problem. We discuss the complete identify algorithm in section four and prove the sufficiency of the three inference rules in section five. Conclusions are included in section six.

---

[1] The notations $P(s|do(t))$ and $P(s|\hat{t})$ are used in [Pearl, 2000], and the notation $P_t(s)$ is used in [Tian and Pearl, 2002b, Tian and Pearl, 2003].

## 2 Definitions and Notations

A *causal Bayesian network* consists of a DAG $G$ over a set of variables $V = \{V_1, \ldots, V_n\}$, called a *causal graph*, and a probability distribution on $V$. The interpretation of this kind of model consists of two parts. One is the probabilistic interpretation, which says that each variable in the graph is independent of all its non-descendants given its direct parents; the other is the causal interpretation, which says that the directed edges in $G$ represent causal influences between the corresponding variables [Pearl, 2000, Lauritzen, 2001].

We use $V(G)$ to show that $V$ is the variable set of graph $G$. If it is clear in the context, we also use $V$ directly. Capital characters, like $V$, are used for variable sets; the lower characters, like $v$, stand for the instances of variable set $V$. Capital character like $X$, $Y$ and $V_i$ are also used for single variable, and their values can be $x$, $y$ and $v_i$. Normally, we use $F(V)$ to denote a function on variable set $V$. An instance of this function is denoted as $F(V)(V = v)$, or $F(V)(v)$, or just $F(v)$. Because all the variables are in the causal graph, we sometimes use node and node set instead of variable and variable set. We use $Pa(V_i)$ to denote parent node set of node $V_i$ in graph $G$ and $pa(V_i)$ as an instance of variable set $Pa(V_i)$. $Ch(V_i)$ is $V_i$'s children node set; $ch(V_i)$ is an instance of $Ch(V_i)$.

Based on the probabilistic interpretation of causal Bayesian network, we have that the joint probability function $P(v) = P(v_1, \ldots, v_n)$ can be factorized as

$$P(v) = \prod_{V_i \in V} P(v_i | pa(V_i)) \quad (1)$$

The causal interpretation of Markovian model enables us to predict intervention effects. Here, intervention means some kind of modification of factors in product (1). The simplest kind of intervention is fixing a subset $T \subseteq V$ of variables to some constants $t$, denoted by $do(T = t)$ or just $do(t)$, and then the post-intervention distribution

$$P_T(V)(T = t, V = v) = P_t(v) \quad (2)$$

is given by:

$$P_t(v) = \begin{cases} \prod_{V_i \in V \setminus T} P(v_i | pa(V_i)) & v \text{ consistent with } t \\ 0 & v \text{ inconsistent with } t \end{cases} \quad (3)$$

We note explicitly that the post-intervention distribution $P_t(v)$ is a probability distribution.

When all the variables in $V$ are observable, since all $P(v_i | pa_i)$ can be estimated from nonexperimental data, as just indicated, all causal effects are computable. But when some variables in $V$ are unobservable, things are much more complex.

Let $N$ and $U$ stand for the sets of observable and unobservable variables in graph $G$ respectively, that is $V = N \cup U$. The observed probability distribution $P(n) = P(N = n)$, is a mixture of products:

$$P(n) = \sum_U \prod_{V_i \in N} P(v_i | pa(V_i)) \prod_{V_j \in U} P(v_j | pa(V_j)) \quad (4)$$

The post-intervention distribution $P_t(n) = P_{T=t}(N = n)$ is defined as:

$$P_t(n) = \begin{cases} \sum_U \prod_{V_i \in N \setminus T} P(v_i | pa(V_i)) \times \\ \prod_{V_j \in U} P(v_j | pa(V_j)) \\ \quad n \text{ consistent with } t \\ 0 \quad n \text{ inconsistent with } t \end{cases} \quad (5)$$

Sometimes what we want to know is not the post-intervention distribution for the whole $N$, but the post-intervention distribution $P_t(s)$ of an observable variable subset $S \subset N$, For those two observable variable set $S$ and $T$, $P_t(s) = P_{T=t}(S = s)$ is given by:

$$P_t(s) = \begin{cases} \sum_{V_i \in (N \setminus S) \setminus T} \sum_U \prod_{V_i \in N \setminus T} P(v_i | pa(V_i)) \\ \prod_{V_j \in U} P(v_j | pa(V_j)) \\ \quad s \text{ consistent with } t \\ 0 \quad s \text{ inconsistent with } t \end{cases} \quad (6)$$

The identifiability question is defined as whether the causal effect $P_T(S)$, that is all $P_t(s)$ given by (6), can be determined uniquely from the distribution $P(N = n)$ given by (4), and thus independent of the unknown quantities $P(v_i | pa(V_i))$s, where $V_i \in U$ or there are some $V_j \in Pa(V_i), V_j \in U$.

We give out a formal definition of *identifiability* below, which follows [Tian and Pearl, 2003].

A Markovian model consists of four elements

$$M = < N, U, G_{N \cup U}, P(v_i | pa(V_i)) >$$

where, (i) $N = \{N_1, \ldots, N_m\}$ is a set of observable variables; (ii) $U = \{U_1, \ldots, U_n\}$ is a set of unobservable variables; (iii) $G$ is a directed acyclic graph with nodes corresponding to the elements of $V = N \cup U$; and (vi) $P(v_i|pa(V_i))$, $i = 1, \ldots, m+n$, is the conditional probability of variable $V_i \in V$ given its parents $Pa(V_i))$ in $G$.

**Definition 1** The causal effect of a set of variables $T$ on a disjoint set of variables $S$ is said to be identifiable from a graph $G$ if all the quantities $P_t(s)$ can be computed uniquely from any positive probability of the observed variables — that is , if $P_t^{M_1}(s) = P_t^{M_2}(s)$ for every pair of models $M_1$ and $M_2$ with $P^{M_1}(n) = P^{M_2}(n) > 0$ and $G(M_1) = G(M_2)$.

This definition means that, given the causal graph $G$, the quantity $P_t(s)$ can be determined from the observed distribution $P(n)$ alone.

Normally, when we talk about $S$ and $T$, we think they are both observable variable subsets of $N$ and mutually disjoint. So, $s$ is always consistent with $t$ in Equation 6.

We are sometimes interested in the causal effect on a set of observable variables $S$ due to all other observable variables. In this case, keeping the convention that $N$ stands for the set of all observable variables, $T = N \backslash S$. For convenience and for uniformity with [Tian and Pearl, 2002b], we define

$$Q[S] = P_{N \backslash S}(S) \qquad (7)$$

and interpret this equation as stating that $Q[S]$ is the causal effect of $N \backslash S$ on $S$.

We define the *c-component relation* on the unobserved variable set $U$ of graph $G$ as follow: for any $U_1 \in U$ and $U_2 \in U$, they are related under the c-component relation if and only if at least one of conditions below is satisfied:

(i) there is an edge between $U_1$ and $U_2$,

(ii) $U_1$ and $U_2$ are both parents of the same observable node,

(iii) both $U_1$ and $U_2$ are in the c-component relation with respect to another node $U_3 \in U$.

Observe that the c-component relation in $U$ is reflexive, symmetric and transitive, so it defines a partition of $U$. Based on this relationship, we can therefore divide $U$ into disjoint and mutually exclusive c-component related parts.

A *c-component* of variable set $V$ on graph $G$ consists of all the unobservable variables belonging to the same c-component related part of $U$ and all observable variables that have an unobservable parent which is a member of that c-component. According to the definition of c-component relation, it is clear that an observable node can only appear in one c-component. If an observable node has no unobservable parent, then itself is a c-component on $V$. Therefore, the c-components form a partition on all of the variables.

We conclude this section by giving several simple graphical definitions that will be needed later.

For a given set of variables $C$, we define the *directed unobservable parent set* $DUP(C)$. A node $V$ belongs to $DUP(C)$ if and only if both of these two conditions are satisfied: i) $V$ is an unobservable node; ii) there is a directed path from $V$ to an element of $C$, and all the internal nodes on that path are unobservable nodes.

For a given observable variable set $C \subseteq N$, let $G_C$ denote the subgraph of $G$ composed only of variables in $C \cup DUP(C)$ and all the links between variable pairs in $C \cup DUP(C)$. Let $An(C)$ be the union of $C$ and the set of ancestors of the variables in $C$, and let $De(C)$ be the union of $C$ and the set of descendents of the variables in $C$. An observable variable set $S \subseteq N$ in graph $G$ is called an *ancestral set* if it contains all its own observed ancestors, i.e., $S = An(S) \cap N$.

## 3 Graphical Criteria

In general, to solve the identifiability problem graphically, there are two things we need to know. The first is a set of inference rules, which can transfer causal effect expressions to equivalent expressions. The second is a sound and complete algorithm based on those rules. Here complete means that, for any causal effect question, we can use this algorithm to answer it, either by generating a final expression just involving ordinary conditional probabilities, which is assessable by empirical observation, or by reporting that the effect is unidentifiable.

Using causal Bayesian network, Pearl gives two graphical criteria to check identifiability, and called them *back-door criterion* and *front-door criterion* [Pearl, 1993, Pearl, 1995]. Following these results, in [Pearl, 1995] and also in his book *Causality* [Pearl, 2000], Pearl proposes three inference rules (the *do-calculus rules*) that allow transformations between sentences concerning interventions and observations. The aim of such rules is to lead to a calculus of interventions and observations on causal Bayesian networks, so that, whenever possible, sentences that involve interventions and observations may be transformed into sentences that involve only observations. Pearl proves that the three *do*-calculus rules are sound and conjectures that they are sufficient. We present these three inference rules below. We begin by reviewing several definitions from [Pearl, 2000].

Let $X$, $Y$, $Z$ be arbitrary disjoint sets of nodes in a causal graph $G$. We denote by $G_{\overline{X}}$ the graph obtained by deleting from $G$ all arrows pointing to nodes in $X$. Likewise, we denote by $G_{\underline{X}}$ the graph obtained by deleting from $G$ all arrows emerging from nodes in $X$. To represent the deletion of both incoming and outgoing arrows, we use the notation $G_{\overline{X}\underline{Z}}$.

The expression $P(y|\hat{x}, z) \equiv P(y, z|\hat{x})/P(z|\hat{x})$ stands for the probability of $Y = y$ given that $X$ is held constant at $x$ and that (under this condition) $Z = z$ is observed. Our notation for this probability is $P_x(y|z)$. In this section, we use Pearl's notation in [Pearl, 2000] when quoting his results directly.

Here are the three inference rules of proposed calculus. Proofs of soundness can be found in [Pearl, 1995].

**Theorem** (*Rules of Do-Calculus*) [Pearl, 2000] Let $G$ be the directed acyclic graph associated with a causal model, and let $P(\dot{)}$ stand for the probability distribution induced by that model. For any disjoint subsets of variables $X, Y, Z$, and $W$ we have the following rules.

Rule 1 (Insertion/deletion of observations)

$$P(y|\hat{x}, z, w) = P(y|\hat{x}, w) \ if (Y \perp Z | X, W)_{G_{\overline{X}}} \quad (8)$$

Rule 2 (Action/observation exchange)

$$P(y|\hat{x}, \hat{z}, w) = P(y|\hat{x}, z, w) \ if (Y \perp Z | X, W)_{G_{\overline{X}\underline{Z}}} \quad (9)$$

Rule 3 (Insertion/deletion of actions)

$$P(y|\hat{x}, \hat{z}, w) = P(y|\hat{x}, w) \ if (Y \perp Z | X, W)_{G_{\overline{X}, \overline{Z(W)}}} \quad (10)$$

where $Z(W)$ is the set of $Z$-nodes that are not ancestors of any $W$-node in $G_{\overline{X}}$.

In [Pearl, 2000], the author shows that both his backdoor and front-door criteria can be obtained from these three rules. In [Galles and Pearl, 1995], the authors give out a graphical criterion to identify the causal effect between a singleton variable $X$ and a set of variables $Y$. Their algorithm works in time polynomial on the number of variables in the graph. This result is also showed in [Pearl, 2000][2]. In [Pearl and Robins, 1995], the authors extend the results of [Galles and Pearl, 1995] to the case where $T$ stands for a compound action, consisting of several atomic interventions that are implemented either concurrently or sequentially. They establish a graphical criterion for recognizing when the effect of $X$ on $Y$ is identifiable and, in case the diagram satisfies this criterion, they provide a closed-form expression for the distribution of an outcome variable $S$ under the plan defined by the compound action setting $T = t$. Following Pearl and Robins' work, [Kuroki and Miyakawa, 1999] present an extension of the front door criterion.

All the criteria cited above are based the three inference rules. Therefore, the proof of sufficiency of the three rules that is provided in this paper paves the road for proofs of sufficiency of other graphical algorithms in this area.

---

[2]We explicitly note that the algorithm given in Section 4.3.1 of [Pearl, 2000], while inspired by the do-calculus, is not complete, as shown in [Tian and Pearl, 2003].

## 4 A Sound and Complete Identification Algorithm

In this section we present a complete identification algorithm.

For a given model (causal Bayesian network) with graph $G$, We begin with removing all unobservable nodes that have no observable descendants. From the definitions in section two, it is easy to prove that this transformation does not change the identifiability of the model [Huang and Valtorta, 2006a].

Below are two lemmas proved by Tian and Pearl in [Tian and Pearl, 2002b].

**Lemma 1** *Let $W \subseteq C \subseteq N$. If $W$ is an ancestral set in $G_C$, then*

$$\sum_{V_i \in C \setminus W} Q[C] = Q[W] \quad (11)$$

**Lemma 2** *Let $H \subseteq N$, and we have c-components $H'_1, \ldots, H'_n$ in the sub graph $G_H$, $H_i = H'_i \cap H$, $1 \leqslant i \leqslant n$, then*

*(i) $Q[H]$ can be decomposed as*

$$Q[H] = \prod_{i=1}^{n} Q[H_i] \quad (12)$$

*(ii) Each $Q[H_i]$ is computable from $Q[H]$, in the following way. Let $k$ be the number of variables in $H$, and let a topological order of variables in $H$ be $V_{h_1} < \ldots < V_{h_k}$ in $G_H$. Let $H^{(j)} = \{V_{h_1}, \ldots, V_{h_j}\}$ be the set of variables in $H$ ordered before $V_{h_j}$ (including $V_{h_j}$), $j = 1, \ldots, k$, and $H^{(0)} = \phi$. Then each $Q[H_i], i = 1, \ldots, n$, is given by*

$$Q[H_i] = \prod_{\{j|V_{h_j} \in H_i\}} \frac{Q[H^{(j)}]}{Q[H^{(j-1)}]} \quad (13)$$

*where each $Q[H^{(j)}]$, $j = 0, 1, \ldots, k$, is given by*

$$Q[H^{(j)}] = \sum_{h \setminus h^{(j)}} Q[H] \quad (14)$$

Assume that $N(G)$ is partitioned into $N_1, \ldots, N_k$ in $G$, where each $N_i$ belongs to a c-component, and that we have c-components $S'_1, \ldots, S'_l$ in graph $G_S$, $S_j = S'_j \cap S$, $1 \leqslant j \leqslant l$. Based on lemma 2, for any model on graph $G$, we have

$$Q[S] = \prod_{j=1}^{l} Q[S_j] \quad (15)$$

Because each $S_j, j = 1, \ldots, l$, which is a c-component in $G_S$, is a subgraph of $G$, it must be included in one $N_j$, $N_j \in \{N_1, \ldots, N_k\}$.

The rest of this section is devoted to three algorithms.

**Algorithm Computing** $Q[S]$

INPUT: $S \subseteq N$

OUTPUT: Expression for $Q[S]$ or FAIL

Let $N(G)$ be partitioned into $N_1, \ldots, N_k$, each of them belonging to a c-components in $G$, and $S$ be partitioned into $S_1, \ldots, S_l$, each of them belonging to a c-components in $G_S$, and $S_j \subseteq N_j$. We can

i) Compute each $Q[N_j]$ with lemma 2

ii) Compute each $Q[S_j]$ with Identify algorithm below with $C = S_j, T = N_j, Q = Q[N_j]$

iii) If in ii), we get Fail as return value of Identify algorithm of any $S_j$, then $Q[S]$ is unidentifiable in graph $G$; else $Q[S]$ is identifiable and equal to $\prod_{j=1}^{l} Q[S_j]$ (by lemma 2)

**Algorithm Identify** ($C$,$T$,$Q$)

INPUT: $C \subseteq T \subseteq N$, $Q = Q[T]$, $G_T$ and $G_C$ are both composed of one single c-component

OUTPUT: Expression for $Q[C]$ in terms of $Q$ or FAIL

Let $A = An(C)_{G_T} \cap T$

i) If $A = C$, output $Q[C]$, which is equal to $\sum_{T \setminus C} Q[T]$ by lemma 1

ii) Else if $A = T$, output FAIL

iii) Else (if $C \subset A \subset T$)

1. Compute $Q[A] = \sum_{T \setminus A} Q[T]$ with lemma 1

2. Assume that in $G_A$, $C$ is contained in a c-component $T_1'$, $T_1 = T_1' \cap A$.

3. Compute $Q[T_1]$ from $Q[A]$ with lemma 2

4. Output Identify($C$,$T_1$,$Q[T_1]$)

To compute $P_T(S)$, we can rewrite it as:

$$P_t(s) = \sum_{N \setminus (T \cup S)} P_t(n \setminus t) = \sum_{N \setminus (T \cup S)} Q[N \setminus T] \quad (16)$$

Let $D = An(S)_{G_{N \setminus T}} \cap N$. $D$ is an ancestral set in graph $G_{N \setminus T}$. Lemma 1 allows us to conclude that $\sum_{N \setminus (T \cup D)} Q[N \setminus T] = Q[D]$. Therefore, we have:

$$P_t(s) = \sum_{D \setminus S} \sum_{N \setminus (T \cup D)} Q[N \setminus T] = \sum_{D \setminus S} Q[D] \quad (17)$$

**Algorithm Computing** $P_T(S)$

INPUT: two disjoint observable variable sets $S, T \subset N$

OUTPUT: the expression for $P_T(S)$ or FAIL

1. Let $D = An(S)_{G_{N \setminus T}} \cap N$

2. Use the Computing $Q[S]$ algorithm to compute $Q[D]$

3. If the algorithm returns FAIL, then output FAIL

4. Else, output $P_T(S) = \sum_{D \setminus S} Q[D]$

The authors, in [Huang and Valtorta, 2006b], prove that:

**Theorem 1** *The above algorithm for computing $P_T(S)$ is sound and complete.*

Note that the soundness of the algorithms above can be obtained from lemma 1, 2 and standard probability manipulations. We will exploit this property in the next section.

## 5 Completeness of the Three Inference Rules of Pearl's Do-Calculus

In this section, we prove the three inference rules are complete. As already mentioned, the soundness of the three rules is proved in [Pearl, 1995]. Here we just need to prove their sufficiency.

Note that the sound and complete algorithm for computing $P_T(S)$ in the last section is obtained by using lemma 1 and lemma 2. If we can show that lemma 1 and lemma 2 can be obtained through just using the three inference rules and standard probability manipulations, then the sufficiency of the three rules is proved.

We begin with the following observation:

**Lemma 3** *If any of the three rules can be used on a model with graph $G$, it can also be used on a model that is obtained by removing all unobservable nodes that have no observable descendants.*

*Proof*: This follows from the well-known result that barren nodes can be removed without changing the d-separation relation for the other nodes [Shachter, 1986]. □

We also have:

**Lemma 4** *Rule 1 follows from rule 2 and rule 3.*

*Proof*: Since removing an edge can only d-separate more nodes in a Bayesian network, the conditions for the application of rules 2 and 3 are satisfied if the condition for rule 1 is satisfied. We can replace the application of rule 1 by the application of rule 2 followed by the application of rule 3. In detail, by applying rule 2, we have that $P(y|\hat{x}, z, w) = P(y|\hat{x}, \hat{z}, w)$. By applying rule 3, we have that $P(y|\hat{x}, \hat{z}, w) = P(y|\hat{x}, w)$. So, $P(y|\hat{x}, z, w) = P(y|\hat{x}, w)$, which is the result of applying rule 1. □

We now show that lemma 1 and lemma 2 can be obtained through just using rule 2, rule 3 and standard probability manipulations. For convenience, we define $V_1^m =$

$\{V_1, V_2, \ldots, V_m\}$ and $\hat{V}_1^m = \{\hat{V}_1, \hat{V}_2, \ldots, \hat{V}_m\}$. Here, $V_i$, $1 \leq i \leq m$ can be any variable or variable set; in the case of sets, the comma should be understood as the union operator.

**Lemma 5** *Lemma 1 follows from rule 3.*

*Proof*: We recall the definition of ancestral set. An observable variable set $S \subseteq N$ in graph $G$ is called an *ancestral set* if it contains all its own observed ancestors, i.e., $S = An(S) \cap N$. Because $G$ is a DAG, $G_C$, which include all nodes in $C \cup DUP(C)$, is a DAG also. If $W$ is an ancestral set in $G_C$, then there is a topological order of nodes in $G_C$ that starts with all the nodes in $W$ and continues with the other nodes. If $W = N$, the lemma is trivially true. Otherwise, consider the first node, say $X$, in the topological order just described and that is in $C$ but not in $W$. So, what we need to prove is: If $W \subset N$, node $X \in N \backslash W$, $C = W \cup \{X\}$, and $W$ is an ancestral set in $G_C$, then

$$\sum_X Q[C] = Q[W] \quad (18)$$

Recall that for any $S \subset N$, by definition $Q[S] = P_{N \backslash S}(S)$. So, equation 18 can be rewritten as

$$\sum_X P_{N \backslash C}(C) = P_{N \backslash W}(W) \quad (19)$$

Assume $W = \{X_1, \ldots, X_k\} = X_1^k$, $Y = N \backslash (W \cup \{X\})$, where, unlike $X$, $Y$ is a variable set here. Equation 19 becomes

$$\sum_X P_Y(X_1^k, X) = P_{Y,X}(X_1^k) \quad (20)$$

Using the graphical language Pearl used, what we want to prove is

$$\sum_X P(X_1^k, X|\hat{Y}) = P(X_1^k|\hat{Y}, \hat{X}) \quad (21)$$

We know that

$$\sum_X P(X_1^k, X|\hat{Y}) = \sum_X P(X|X_1^k, \hat{Y}) P(X_1^k|\hat{Y})$$
$$= (\sum_X P(X|X_1^k, \hat{Y})) P(X_1^k|\hat{Y}) \quad (22)$$

Now, we use rule 3 of the do-calculus. Note that we can apply this rule, because for $P(X_1^k|\hat{Y})$, we have

$$(\{X_1^k\} \perp X|Y)_{G_{\overline{Y}, \overline{X}}} \quad (23)$$

This is because in graph $G_{\overline{Y}, \overline{X}}$, if there is a d-connected path from $X$ to a node $X_i$, $1 \leq i \leq k$, that path could not include any node in $Y$, because $Y$ nodes can only be divergent nodes and $Y$ is given. If that path just goes from $X$ through some unobservable nodes to $X_i$, it would mean that topologically $X$ is before $X_i$ in graph $G_C$, which could not be true. So, that kind of path does not exist.

Using rule 3 we obtain

$$P(X_1^k|\hat{Y}) = P(X_1^k|\hat{Y}, \hat{X}) \quad (24)$$

So, what we need to prove is just

$$\sum_X P(X|X_1^k, \hat{Y})) = 1 \quad (25)$$

This is obvious, since $P_y$ is a probability distribution. □

**Lemma 6** *Lemma 2 follows from rule 2 and rules 3.*

*Proof*: When $H$ just includes one observable variable, (i) and (ii) in lemma 2 are clearly true.

Assume (i) and (ii) are still true for any observable variable set $H \subset N$, where the size of $H$ is less than or equal to integer $k$.

Consider an arbitrary observable variable set $E$ of size $k+1$, and assume variable $X \in E$ is topologically after all the variables in $H = E \backslash \{X\}$ in graph $G$. Assume $H \cup DUP(H)$ can be divided into c-component $H_1', \ldots, H_n'$ in graph $G_H$, and $H_i = H_i' \cap H, 1 \leq i \leq n$. Let $Y = N \backslash E$. Also assume $X$ and $H_1, H_2, \ldots, H_m$, $0 \leq m \leq n$ construct a c-component in graph $G_E$. (If $m = 0$, then $X$ is a c-component by itself.)

Since the size of $H$ is $k$, we have the inductive hypothesis

$$Q[H] = \prod_{i=1}^n Q[H_i] \quad (26)$$

which means:

$$P(H|\hat{Y}, \hat{X}) = \prod_{i=1}^n P(H_i|\hat{H}_1^{i-1}, \hat{H}_{i+1}^n, \hat{Y}, \hat{X}) \quad (27)$$

What we want to prove is that (i) and (ii) are still true for $E$.

For (i), we want to prove

$$P(H, X|\hat{Y}) = P(H_1^m, X|\hat{H}_{m+1}^n, \hat{Y}) \times \\ \prod_{j=m+1}^n P(H_j|\hat{H}_1^{j-1}, \hat{H}_{j+1}^n, \hat{Y}, \hat{X}) \quad (28)$$

Note that we have

$$P(H, X|\hat{Y}) = P(X|H, \hat{Y}) P(H|\hat{Y}) \quad (29)$$

We know that, in graph $G_{\overline{Y}, \overline{X}}$, $(H \perp \{X\}|Y)$. This is because if there is a d-connected path from $X$ to any node of $H$ in $G_{\overline{Y}, \overline{X}}$, that path could not include any node in $Y$, since $Y$ nodes can only be divergent nodes. Then that path must go from $X$ to one node of $H$, and this is impossible because we assume $X$ is topologically after all nodes in $H$.

Based on rule 3, we have

$$P(H|\hat{Y}) = P(H|\hat{X}, \hat{Y}) \tag{30}$$

Then 29 can be rewritten as

$$\begin{aligned} P(H, X|\hat{Y}) &= P(X|H, \hat{Y})P(H|\hat{X}, \hat{Y}) = \\ &P(X|H, \hat{Y}) \times \prod_{j=1}^{m} P(H_j|\hat{H}_1^{j-1}, \hat{H}_{j+1}^n, \hat{Y}, \hat{X}) \times \\ &\prod_{j=m+1}^{n} P(H_j|\hat{H}_1^{j-1}, \hat{H}_{j+1}^n, \hat{Y}, \hat{X}) \end{aligned} \tag{31}$$

Based on the induction assumption, we have

$$\prod_{j=1}^{m} P(H_j|\hat{H}_1^{j-1}, \hat{H}_{j+1}^n, \hat{Y}, \hat{X}) = P(H_1^m|\hat{H}_{m+1}^n, \hat{Y}, \hat{X}) \tag{32}$$

Just as before, we know that in graph $G_{\overline{Y, H_{m+1}^n}, \overline{X}}$, $(\{H_1^m\} \perp \{X\}|\{Y, H_{m+1}^n\})$. This is because if there is a d-connected path from $X$ to any node of $H_{m+1}^n$ in $G_{\overline{Y, H_{m+1}^n}, \overline{X}}$, that path could not go through any node in $Y$, for all $Y$ nodes can only be divergent nodes. Then that path must go from $X$ to one node of $H$. This is impossible because we assume $X$ is topologically after all nodes in $H$.

With rule 3, we have

$$P(H_1^m|\hat{H}_{m+1}^n, \hat{Y}, \hat{X}) = P(H_1^m|\hat{H}_{m+1}^n, \hat{Y}) \tag{33}$$

We now show that $(X \perp \{H_{m+1}^n\}|Y, H_1^m)$ in graph $G_{\overline{Y}, \underline{H_{m+1}^n}}$. If there is a d-connected path from $X$ to a node $Z$ in $\{H_{m+1}^n\}$ in $G_{\overline{Y}, \underline{H_{m+1}^n}}$, that path could not go through any node in $Y$; assume $\overline{Z'}$ is the nearest observable node to $Z$ on that path, $Z' \in \{H_1^m\}$. If there are some unobservable nodes on that path between $Z$ and $Z'$, then $Z$ and $Z'$ belong to the some c-component, (because $Z'$ can only be a convergent node on that path and the path gets into $Z$), which is impossible; a link from $Z$ to $Z'$ is impossible because all links exiting from $Z$ are removed, and a link from $Z'$ to $Z$ would not open the connection between $Z'$ and $Z$, because $Z'$ is known. So, $Z'$ does not exist. If there are some unobservable nodes between $X$ and $Z$, then $X$ and $Z$ belong to the same c-component in graph $G_E$ (because there must be a divergent unobservable node path between them, otherwise, $X$ is topologically before $Z$), but this is impossible because $Z$ is in $\{H_{m+1}^n\}$. A link from $X$ to $Z$ is also impossible, because we assume $X$ is topologically after all nodes in $H$. So, $X$ and $Z$ are d-separated.

Based on rule 2, we have

$$\begin{aligned} P(X|H, \hat{Y}) &= P(X|H_1^m, H_{m+1}^n, \hat{Y}) = \\ &P(X|H_1^m, \hat{H}_{m+1}^n, \hat{Y}) \end{aligned} \tag{34}$$

Putting 31, 32, 33, and 34 together, we have:

$$\begin{aligned} P(H, X|\hat{Y}) &= \\ P(X|H, \hat{Y}) &\prod_{j=1}^{m} P(H_j|\hat{H}_1^{j-1}, \hat{H}_{j+1}^n, \hat{Y}, \hat{X}) \times \\ \prod_{j=m+1}^{n} &P(H_j|\hat{H}_1^{j-1}, \hat{H}_{j+1}^n, \hat{Y}, \hat{X}) = \\ P(X|H_1^m, &\hat{H}_{m+1}^n, \hat{Y}) \times P(H_1^m|\hat{H}_{m+1}^n, \hat{Y}) \times \\ \prod_{j=m+1}^{n} &P(H_j|\hat{H}_1^{j-1}, \hat{H}_{j+1}^n, \hat{Y}, \hat{X}) = \\ P(H_1^m, X|&\hat{H}_{m+1}^n, \hat{Y}) \times \\ \prod_{j=m+1}^{n} &P(H_j|\hat{H}_1^{j-1}, \hat{H}_{j+1}^n, \hat{Y}, \hat{X}) \end{aligned} \tag{35}$$

Now let us consider the second part of this lemma.

From lemma 1, we have

$$Q[H] = \sum_X Q(H, X) = \sum_X Q[E] \tag{36}$$

Our inductive assumption is that $H$ satisfies (ii), where $H^{(j)} = \{V_{h_1}, \ldots, V_{h_j}\}$ is the set of variables in $H$ ordered before $V_{h_j}$ (including $V_{h_j}$). Then each $Q[H_i], i = 1, \ldots, n$, is given by

$$Q[H_i] = \prod_{\{j|V_{h_j} \in H_i\}} \frac{Q[H^{(j)}]}{Q[H^{(j-1)}]} \tag{37}$$

where each $Q[H^{(j)}], j = 0, 1, \ldots, k$, is given by

$$Q[H^{(j)}] = \sum_{h \setminus h^{(j)}} Q[H] \tag{38}$$

From equation 36, we have

$$Q[E^{(j)}] = \sum_{\{h,x\} \setminus h^{(j)}} Q[E] = \sum_{h \setminus h^{(j)}} Q[H] = Q[H^{(j)}] \tag{39}$$

From (i) we have

$$Q[E] = Q[E^{(k+1)}] = Q[H_1^m, X] \prod_{i=m+1}^{n} Q[H_i] \tag{40}$$

We have

$$Q[H_1^m, X] = Q[E^{(k+1)}] / \prod_{i=m+1}^{n} Q[H_i] \tag{41}$$

The chain decomposition allows us to write

$$Q[E^{k+1}] = \prod_{j=0}^{k+1} Q[E^{(j)}] / Q[E^{(j-1)}] \tag{42}$$

and for each $m+1 \leqslant i \leqslant n$,

$$Q[H_i] = \prod_{\{j|V_{h_j} \in H_i\}} \frac{Q[H^{(j)}]}{Q[H^{(j-1)}]} = \prod_{\{j|V_{h_j} \in H_i\}} \frac{Q[E^{(j)}]}{Q[E^{(j-1)}]} \tag{43}$$

So, equation 41 can be rewritten as

$$Q[H_1^m, X] = Q[E^{(k+1)}] / \prod_{i=m+1}^{n} Q[H_i] = \prod_{\{j|V_{h_j} \in H_1^m, X\}} \frac{Q[E^{(j)}]}{Q[E^{(j-1)}]} \quad (44)$$

□

Putting the lemmas in this section together, we have

**Theorem 2** *The three inference rules, together with standard probability manipulations, are complete for determining identifiability of all effects of $P_T(S)$.*

Theorem 2 confirms Pearl's conjecture that the three rules are sufficient.

## 6 Conclusion

In this paper, we prove that the do-calculus method of [Pearl, 1995] is complete, in the sense that, if a causal effect is identifiable, there exists a sequence of applications of the rules of the do-calculus that transforms the causal effect formula into one that only includes observational quantities.

In fact the constructive proofs in the fifth section do not just show us those rules are complete, but they also provide a formal recursive algorithms to do calculation with rule 2 and 3 when we need lemma 1 or 2. Together with the algorithm we gave in section four, we obtain a formal procedure to solve the identifiability problem and compute causal effects with graphical rules 2 and 3. Clearly, this procedure is complete too. We are not claiming that the procedure just outlined is guided by the structure of the causal graph in a way that would be easy to understand for a causal modeler: this remains an issue to be studied further.

## References

[Galles and Pearl, 1995] Galles, D. and Pearl, J. (1995). Testing identifiability of causal effects. In *Proceedings of UAI-95*, pages 185–195.

[Huang and Valtorta, 2006a] Huang, Y. and Valtorta, M. (2006a). On the completeness of an identifiability algorithm for semi-Markovian models. Technical report, University of South Carolina Department of Computer Science. Available at http://www.cse.sc.edu/ mgv/reports/tr2006-001.pdf.

[Huang and Valtorta, 2006b] Huang, Y. and Valtorta, M. (2006b). A study of identifiability in causal Bayesian network. Technical report, University of South Carolina Department of Computer Science. Available at http://www.cse.sc.edu/ mgv/reports/tr2006-002.pdf.

[Kuroki and Miyakawa, 1999] Kuroki, M. and Miyakawa, M. (1999). Identifiability criteria for causal effects of joint interventions. *Journal of the Japan Statistical Society*, 29(2):105–117.

[Lauritzen, 2001] Lauritzen, S. (2001). Causal inference from graphical models. In Barndorff-Nielsen, O. and Klueppelberg, C., editors, *Complex Stochastic Systems*, pages 63–107. Chapman and Hall, London.

[Pearl, 1993] Pearl, J. (1993). Graphical models, causality, and intervention. comments on: 'Linear Dependencies Represented by Chain Graphs' by D. Cox and N. Wermuth, and 'Bayesian Analysis in Expert Systems' by D.J. Spiegelhalter, A.P. Dawid, S.L. Lauritzen, and R.G. Cowell. In Statistical Science, Vol. 8, 266-269.

[Pearl, 1995] Pearl, J. (1995). Causal diagrams for empirical research. *Biometrika*, 82:669–710.

[Pearl, 2000] Pearl, J. (2000). *Causality: Models, Reasoning, and Inference*. Cambridge University Press, New York.

[Pearl and Robins, 1995] Pearl, J. and Robins, J. M. (1995). Probabilistic evaluation of sequential plans from causal models with hidden variables. In *Proceedings of UAI-95*, pages 444–453.

[Robins, 1997] Robins, J. M. (1997). Causal inference from complex longitudinal data. In Berkane, M., editor, *Latent Variable Modeling with Applications to Causality, Volume 120 of Lecture Notes in Statistics*, pages 69–117. SpringerVerlag, New York.

[Shachter, 1986] Shachter, R. D. (1986). Evaluating influence diagrams. *Operations Research*, 34(6):871–882.

[Shpitser and Pearl, 2006] Shpitser, I. and Pearl, J. (2006). Identification of joint interventional distributions in recursive semi-Markovian causal models. Technical report, Cognitive Systems Laboratory, University of California at Los Angeles. Available at http://ftp.cs.ucla.edu/pub/stat_ser/r327.pdf.

[Tian and Pearl, 2002a] Tian, J. and Pearl, J. (2002a). A general identification condition for causal effects. In *Proceedings of the Eighteenth National Conference on Artificial Intelligence (AAAI-02)*, pages 567–573.

[Tian and Pearl, 2002b] Tian, J. and Pearl, J. (2002b). On the testable implications of causal models with hidden variables. In *Proceedings of UAI-02*, pages 519–527.

[Tian and Pearl, 2003] Tian, J. and Pearl, J. (2003). On the identification of causal effects, Technical report 290-L. Technical report, Cognitive Systems Laboratory, University of California at Los Angeles. Extended version available at http://www.cs.iastate.edu/ jtian/r290-L.pdf.